# Neural Networks from Biological to Artificial and Vice Versa

Abdullatif BABA

*Kuwait College of Science and Technology, Computer Science and Engineering Department, Kuwait*
*University of Turkish Aeronautical Association, Computer Engineering Department, Ankara, Turkey*

**ABSTRACT**

In this paper, we examine how deep learning can be utilized to investigate neural health and the difficulties in interpreting neurological analyses within algorithmic models. The key contribution of this paper is the investigation of the impact of a dead neuron on the performance of artificial neural networks (ANNs). Therefore, we conduct several tests using different training algorithms and activation functions to identify the precise influence of the training process on neighboring neurons and the overall performance of the ANN in such cases. The aim is to assess the potential application of the findings in the biological domain, the expected results may have significant implications for the development of effective treatment strategies for neurological disorders. Successive training phases that incorporate visual and acoustic data derived from past social and familial experiences could be suggested to achieve this goal. Finally, we explore the conceptual analogy between the Adam optimizer and the learning process of the brain by delving into the specifics of both systems while acknowledging their fundamental differences.

## 1. Introduction

Artificial Intelligence has been a fruitful area for drawing inspiration from biological processes to develop new smart techniques and optimization algorithms that are crucial for producing rational decisions and achieving desired outcomes. For example, several architectures of ANNs were presented in the last two decades to approximate the brain inference mechanism. On the other hand, Fuzzy logic was created to profit from the concept of uncertainty or the ambiguous assessment of humans when dealing with computing environments. Moreover, the genetic algorithm represents an obvious example of the optimization search method that utilizes successive operators (random selection, crossover, and mutation) to approach the theory of natural evolution. In the same context, some other optimization algorithms were also recently suggested (Gray wolf, Ant colony, and Bee colony) to imitate their swarm behavior in natural life, i.e. their smart hierarchical strategy when moving for hunting or looking for food sources respectively.

Several of the methods previously mentioned have been utilized within the medical domain. Artificial neural networks (ANNs) possess the ability to analyze various forms of data without relying on their linearity. ANNs can also have their parameters adjusted multiple times to achieve a comparable outcome to the symptoms associated with psychological or behavioral conditions Lanillos, Oliva, Philippsen, Yamashita, Nagai and Cheng (2020). Initial research studies have suggested distinct ANN-based models to replicate cognitive disorders, particularly those related to autism Casanova, Gustafsson and Paplinski (2004); Ruppin, Reggia and Horn (1996). Additionally, ANNs have been explored for modeling Schizophrenia, a severe psychiatric disorder Hoffman and McGlashan (1997). Nowadays, deep learning could be suggested as an assertive tool that could be used to investigate the field of neural health. It is important to note that ANNs don't precisely replicate the biological neurons. Practically, artificial neurons have a simplified model of the behavior of biological neurons, and therefore, some adjustments may need to be made to apply ANN techniques to treat biological neuron disorders. The main difficulty here is interpreting conclusions from neurological analyses and studies in acceptable mathematical or algorithmic models; in other words, we need to find the missing link between biological neuroscience and artificial neural networks. A recent study used a deep neural network to predict the progression of Alzheimer's disease based on MRI scans Ding et al. (2020). Another study utilized ANNs to diagnose autism spectrum disorder based on behavioral and EEG data Alotaiby and Malarout (2018). There are also studies that used genetic algorithms and other optimization techniques to optimize the structure and parameters of ANNs for various neurological tasks Verstraeten, Schrauwen, D'Haene, Stroobandt and Campenhout (2006).

In this paper, we introduce a study that investigates the impact of the sudden death of a neuron during the training of an artificial neural network (ANN) on the performance of the neighboring neurons and the overall performance of the ANN. Various scenarios will be tested, considering different training algorithms and activation functions. The study aims to identify the precise influence of the training process, which could potentially be applied in the biological field to improve the functionality of inhibited or damaged neurons in individuals with behavioral disorders. This could be achieved by implementing multiple training phases that incorporate visual and acoustic data gathered from their past social and familial experiences. Finally, in spite of the considerable difference in complexity and information processing, the article delves into exploring the conceptual analogy between the Adam optimizer and the learning mechanism of the brain.

a.baba@kcst.edu.kw; ababa@thk.edu.tr (A. BABA)
ORCID(s): 0000-0001-5165-4205 (A. BABA)

## 2. ANN Vs SNN (The problem statement)

Artificial Neural Networks (ANNs) are a type of machine learning algorithm that inspires the human brain's neural structure. ANNs are particularly useful for tasks that involve large amounts of data, such as image recognition, natural language processing, and speech recognition. They are also capable of learning complex relationships between inputs and outputs, making them well-suited for tasks that involve non-linear mappings. Training an ANN involves adjusting the network's weights and biases in response to a set of training examples. This process is typically done using an optimization algorithm to minimize the difference between the network's output and the desired output for each given input.

### 2.1. ANNs training issues

Backpropagation is frequently utilized as an algorithm to train ANNs, where the neurons could be fired by different types of activation functions; mainly Sigmoid, ReLU, or LeakyReLU; Figure 1. The backpropagation algorithm computes the error between the predicted output and the actual output for each training vector and then uses this error to adjust the weights in the network. Specifically, the algorithm calculates the gradient of the error with respect to each weight in the network and then updates each weight by moving it in the direction that reduces the error. As the training process continues, the weights are expected to converge gradually toward values that minimize the error in the training data. In other words, the weights should become more and more optimized for the specific task the network is being trained to perform.

It is worth noting that the evolution of the weights during training can be influenced by various factors such as the learning rate, the number of training iterations, the structure of the network, and the complexity of the task. Additionally, it is possible for the weights to become stuck in a local minimum, which can prevent them from reaching the global minimum and therefore limit the network's performance.

During the training process, two issues could be encountered:

The first one is called the vanishing gradient problem; where in some cases, the back-propagated gradient errors become vanishingly small, consequently preventing the weights of a given neuron from changing even in small values; that means the neuron itself doesn't profit from the successive training vectors to extract the hidden features from the given data, consequently its influence on the output becomes so small or negligible. In any case, we have to notice that the weights of links inside an ANN are used as long-term memory compared to the biological neuron. One possible solution that could be suggested is to use the ReLU activation function instead of the Sigmoid, as it is faster and does not activate all neurons simultaneously. Another possible solution is to use an optimization algorithm, such as the Adam optimizer Kingma and Ba (2015), which has been shown to perform better than backpropagation in some cases. Additionally, other activation functions such as the hyperbolic tangent (tanh) Glorot, Bordes and Bengio (2011) or the exponential linear unit (ELU) Clevert, Unterthiner and Hochreiter (2015) have also been suggested to address this issue. However, using the ReLU activation function may result in dead neurons that produce zero output in the relevant region, as the gradient for any negative value applied to it is zero and the corresponding weights will not be updated during backpropagation. This issue can be addressed by modifying the activation function to use LeakyReLU, which has a small slope for negative values instead of a flat slope, allowing negative inputs to produce some corresponding values on the output.

Interestingly, behavioral disorders in humans may be caused by inhibited, extra-excited, or damaged neurons, which could correspond to the vanishing gradient problem and dead artificial neurons in ANN. While replacing or modifying the activation function is a feasible solution for ANN, it is more challenging for biological neurons as their activation functions are chemical in nature. Thus, the question of how to replace or modify them remains a significant challenge.

From a technical point of view, reconfigurable FPGA-based implementation could be suggested to model the biological activation function to get sufficient flexibility to track the neuron's evolutionary state during its training process. An alternative approach is to use spiking neural networks (SNNs) Maass (1997a); Wang, Li, Chen and Xu (2020) that are more biologically realistic than traditional ANNs as they incorporate the concept of spikes or action potentials that occur in biological neurons Srinivasa, Cruz-Albrecht, Chakradhar and Cauwenberghs (2016); Moradi, Qiao and Stefanini (2017). They can also simulate the behavior of inhibitory and excitatory neurons, which is important in modeling neural disorders. There are existing tools and frameworks for implementing SNNs on FPGAs, such as SpiNNaker and BrainScaleS Davies, Srinivasa, Lin, Chinya, Cao, Choday and et al. (2018).

### 2.2. Spiking Neural Networks (SNNs)

Spiking Neural Networks (SNNs) are a type of artificial neural network that is inspired by the way that biological neurons communicate through the use of electrical spikes or "action potentials" Maass (1997b). In contrast to traditional neural networks, which typically process continuous-valued inputs and outputs, SNNs operate on time-varying signals, where each neuron is activated by a series of discrete spikes representing the timing and strength of incoming signals Gerstner, Kistler, Naud and Paninski (2014). In SNNs, the information is processed in a more biologically plausible way, where the output of a neuron is determined not only by the magnitude of the input but also by the timing and frequency of the incoming spikes. This allows SNNs to more accurately model the spatiotemporal dynamics of biological neural networks and to handle problems that are difficult for

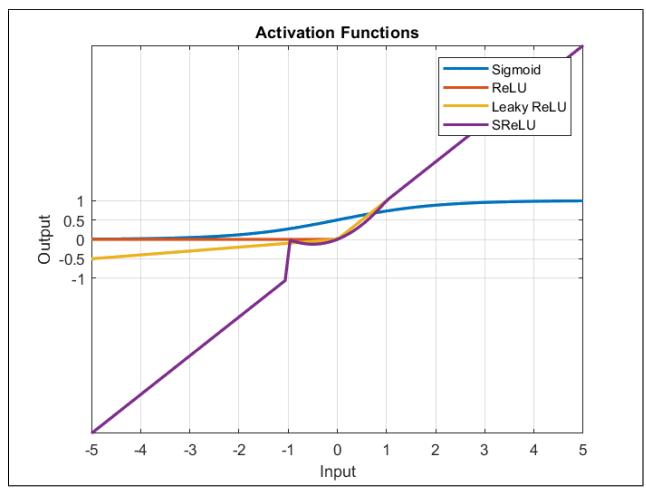

**Figure 1:** The activation functions are illustrated: Sigmoid, ReLU, LeakyReLU, and SReLU

**Table 1**
A Python code clarifying the Spikeing Rectified Linear Unit

```python
def SReLU(x, a, b):
    if x > b:
        return x
    elif x < -a:
        return 0
    else:
        return (x + a) * (b - x) / (b + a)
```

traditional neural networks, such as those involving temporal patterns or event-based data. One important feature of SNNs is their ability to implement a form of time-based computation, where the timing of input signals can be used to perform complex computations Maass (2002). For example, SNNs have been used to perform tasks such as speech recognition, image recognition, and control of robotic systems, where the timing of events is critical to the task at hand Pfeiffer and Brette (2018). SNNs can be trained using a variety of methods, including supervised learning, unsupervised learning, and reinforcement learning, and there are many different architectures and variants of SNNs that have been proposed in the literature Bengio, Courville and Vincent (2015).

There are several algorithms that can be used to train SNNs, but one of the most common is the SpikeProp algorithm, which is an extension of the backpropagation algorithm used in ANNs. It propagates spikes through the network backward from the output to the input layer, similar to how backpropagation propagates error signals in ANNs. Another popular algorithm for training SNNs is the SUR (Synaptic Update Rule) algorithm which is a simple and biologically plausible learning rule that modifies synaptic weights based on the spike timing between the pre-and post-synaptic neurons; i.e. it increases the weight of a connection if the pre-synaptic neuron fires just before the post-synaptic neuron, and decreases it if the pre-synaptic neuron fires just after the post-synaptic neuron. This algorithm aims to strengthen connections that are active at the same time and weaken connections that are not. In addition to these algorithms, there are also unsupervised learning algorithms for training SNNs, such as STDP (Spike-Timing-Dependent Plasticity) that also adjusts the synaptic weights based on the timing of pre-synaptic and post-synaptic spikes, but it does so by using slightly different rules for updating the weights compared to SUR algorithm.

Regarding the activation function, the preferred activation function to use in an SNN depends on the specific task and network architecture. One common activation function used in SNNs is the Sigmoid function, which is also commonly used in ANNs. However, other activation functions, such as the ReLU (Rectified Linear Unit) and its variants, can also be used in SNNs. There are also specialized activation functions designed for SNNs, such as the Spiking Rectified Linear Unit (SReLU) which is a variant of the ReLU that considers the spiking nature of SNNs. SReLU is shown in Figure 1, and given as Python code in Table 1.

## 3. Experiments and analysis

This section assumes training an artificial neural network (ANN) using various algorithms and activation functions to stimulate the neurons in the network. While the ANN is being trained, it is possible for a neuron to lose all relevant weights without any apparent reason. If this happens, we conduct a thorough analysis to understand the expected behavior of the ANN, the affected neuron, and its neighboring neurons. To perform this experiment, we constructed a customized ANN consisting of an input layer with 5 neurons, 3 hidden layers with 10 neurons each, and an output layer with a single neuron, Figure 2. The dataset used for training this network provides information on the power consumption in a specific area over 5 consecutive years. By detecting the hidden patterns within this data, the ANN is capable of predicting the power consumption for the same day of the next year. The dataset was separated into training and testing sets with different and shuffled split ratios that were applied in successive cycles by employing the "KFold" class and the "cross_val_score" function from the "sklearn" Python module. In this study, we are not focusing on the accuracy of the prediction process or adjusting the parameters of the architecture, as these were already discussed in a previous study Baba (2022, 2021). Instead, we are using the previously established and fine-tuned architecture to examine the internal behavior of the ANN when one of its neurons suddenly malfunctions.

Now, let's examine each of these scenarios one by one to deduce the corresponding results:

### 3.1. The first case: (The ANN is trained using the Backpropagation training algorithm. Neurons are fired using the Sigmoid activation function)

When a neuron in an ANN loses its relevant weights, it effectively becomes "dead" and stops contributing to the output of the network. However, the other neurons in the same layer as the lost neuron (the neighbors) compensate for this loss by increasing their own weights. The compensation mechanism occurs during the training process, as the weights of the remaining neurons in the same layer are adjusted to minimize the overall loss of the network. Specifically, the weights of the remaining neurons are adjusted to produce the desired output of the network in response to the input data. In doing so, the network redistributes the contribution of the lost neuron to its neighboring neurons, allowing the network to continue learning to make accurate predictions. The behavior of the neuron with the lost weights itself would depend on the specific implementation of the backpropagation algorithm. If the implementation uses regularization or dropout techniques, then the neuron may continue to contribute to the output of the ANN. Regularization and dropout techniques are methods used in machine learning to prevent the overfitting of a model which usually occurs when a model becomes too complex, resulting in it memorizing the training data rather than learning the underlying patterns that generalize to new data. Regularization involves adding a penalty term to the loss function that the model minimizes during training. This penalty term is a function of the model's parameters and helps to prevent the model from becoming too complex. Dropout is another technique where, during training, a random subset of the model's neurons is deactivated or "dropped out" with a certain probability. This encourages the remaining neurons to learn more robust features and reduces the model's reliance on any particular subset of neurons. Dropout has been shown to be effective in preventing overfitting and improving the generalization performance of deep neural networks.

### 3.2. The second case: (The ANN is trained using the Backpropagation training algorithm. Neurons are fired using the ReLU activation function)

In such a scenario, the behavior of the affected neuron and its neighboring neurons may differ from what is expected with the Sigmoid activation function. In contrast, if a neuron loses its relevant weights, its output will be zero for all inputs. Hence, the lost neuron will not contribute to the ANN's output, and its neighboring neurons will be unable to compensate for the lost contribution since the output of the lost neuron is already zero. Furthermore, with ReLU and depending on the input value, the activation function's gradients are either 0 or 1. If a neuron with zero output has a non-zero gradient, the backpropagation algorithm will be unable to update the weights of the previous layer's neurons, resulting in the vanishing gradient problem. This can make training the ANN difficult or impossible, as the gradients will become extremely small, and the weights will not be updated significantly. Therefore, the loss of relevant weights of a neuron in the ReLU activation function can have a more severe impact on the neuron's behavior and its neighbors in the ANN than with the Sigmoid activation function.

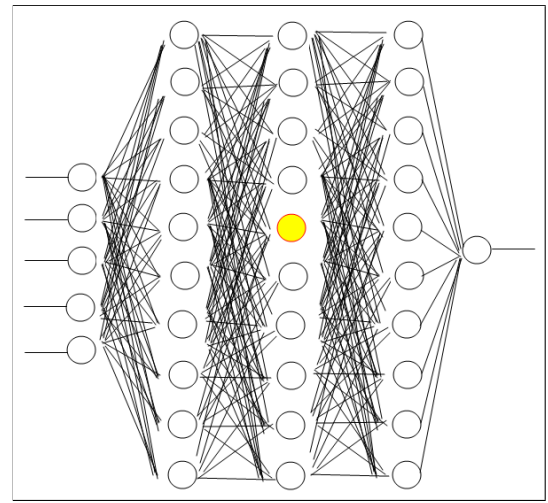

**Figure 2:** The customized ANN consists of an input layer with 5 neurons, 3 hidden layers with 10 neurons each, and an output layer with a single neuron. The yellow neuron is supposed to lose its relevant weights

### 3.3. The third case: (The ANN is trained using the Backpropagation training algorithm. Neurons are fired using the LeakyReLU activation function)

In this case, if a neuron loses its relevant weights, the output of that neuron will not become zero for all inputs; instead, it will be a small non-zero value. Therefore, the behavior of the ANN will change, but the impact will not be as severe as in the case of the ReLU activation function. The neighbors of the lost neuron in the LeakyReLU activation function compensate for the lost contribution to some extent, as the output of the lost neuron will still be non-zero. However, the degree of compensation will depend on the amount of weight loss and the specific topology of the ANN. Moreover, the LeakyReLU gradients are either the leak's slope or 1, depending on the input value. If a neuron with a small non-zero output has a non-zero gradient, then the backpropagation algorithm can still update the neurons' weights in the previous layer. Therefore, the vanishing gradient problem is not as severe as in the case of the ReLU activation function.

### 3.4. The fourth case: (The ANN is trained using the Adam optimizer. Neurons are fired using the Sigmoid activation function)

The Adam optimizer is an optimization algorithm that combines the benefits of both gradient descent and momentum techniques. The algorithm uses adaptive learning rates for each weight parameter to improve convergence speed and stability. Therefore, in the case of weight loss of a neuron, the Adam optimizer will adapt to this change and try to update the affected weights in subsequent iterations to compensate for the lost contribution. If the weight loss is minimal, the impact may not be significant, and the ANN can continue learning without any significant change in behavior.

Interestingly, there are some similarities between the Adam optimizer and the way the human brain learns. One of the key features of the Adam optimizer is the use of adaptive learning rates. In the human brain, a similar process occurs through synaptic plasticity. Synaptic plasticity refers to the ability of the connections between neurons, known as synapses, to change their strength based on the activity of the neurons. This process enables the brain to adapt and learn from new experiences. Moreover, the Adam optimizer uses momentum to help the optimization process. Momentum refers to the tendency of a moving object to continue moving in the same direction. In the human brain, a similar process occurs through the formation of new neural pathways. As a person learns a new skill or task, their brain forms new neural connections that strengthen the existing pathways, making it easier for the person to perform the skill in the future.

### 3.5. The fifth case: (SNN is trained using the backpropagation algorithm. Neurons are fired using the Sigmoid activation function)

If a Spiking Neural Network is trained using backpropagation and one of its neurons loses its relevant weights, the behavior of the SNN, the affected neuron, and its neighbors would be different from that in the case of an ANN. In an SNN, neurons communicate by sending spikes (discrete events) to their neighbors. Therefore, the behavior of a neuron in an SNN depends on the timing and frequency of the spikes it receives from its neighbors. If a neuron loses its relevant weights, it will receive fewer or no spikes from its neighbors, which will reduce its output firing rate. This reduction in firing rate may affect the behavior of the SNN, especially if the lost neuron plays a significant role in the network's computation. Moreover, in an SNN, the output of a neuron is typically binary (spike or no spike), which is different from the continuous output of neurons in an ANN. Therefore, the behaviors of the affected neuron and its neighbors are more complex than in the case of an ANN. The loss of a neuron's relevant weights causes it to stop firing altogether, which affects the firing patterns of its neighbors and the overall computation of the SNN.

### 3.6. The sixth case: (SNN is trained using the SUR (Synaptic Update Rule) algorithm. Neurons are fired using the SReLU activation function)

As mentioned in the last case, the weights of the synapses between neurons are adjusted based on the timing of the pre-and post-synaptic spikes. Therefore, the behavior of a neuron in an SNN depends on the timing and frequency of the spikes it receives from its neighbors, and the loss of its relevant weights could cause it to receive fewer or no spikes, which would reduce its output firing rate. The SReLU activation function is a variant of the ReLU activation function designed for use in SNNs. Like the ReLU, the SReLU function is linear for positive inputs, but it produces a spike (a single output event) instead of continuous output. Therefore, the behavior of an SReLU neuron when it loses its relevant weights could be similar to that of a ReLU neuron in an ANN. The neuron's output firing rate would be reduced, and its neighbors compensate for the lost input by increasing their firing rates.

From the upper-mentioned scenarios, we come to the fundamental conclusion: if a neuron in an artificial neural network loses its relevant weight during backpropagation training, the anticipated behavior of the ANN will be determined by the extent of the weight loss and the stage of training when it occurs. If the lost weights are minor and the training is still in its initial phases, then the effect of the loss may not be substantial, and the ANN may continue to learn with minimal impact on its behavior, while when the lost weights are significant or the training is already in later stages, then the behavior of the ANN may change significantly.

However, as observed in all experiments that are outlined in Table 2, when a neuron suddenly dies, its neighbors on the same layer are activated to make up for the loss. Notably, the closest neurons to the dead neuron bear the greatest responsibility in this process, with their involvement ranging from 60% to 70%, while this proportion decreases significantly with neurons that are farther away.

## 4. From artificial to biological

Biological neural networks are the foundation of the human nervous system and are composed of interconnected neurons that transmit and process information. These networks are responsible for various cognitive functions, including learning, memory, perception, and decision-making. The architecture of biological neural networks consists of individual neurons that are connected through synapses. Neurons receive inputs through dendrites, which are branching extensions that receive signals from other neurons. The signals are then processed in the neuron's cell body or soma. One of the primary activation functions in biological neurons is the action potential, also known as a "spike"; when the accumulated input reaches a certain threshold, the neuron generates an output signal through its axon, which is a long projection that transmits signals to other neurons and the process continues throughout the network.

In this context, the training process refers to the strengthening or weakening of synaptic connections between neurons. This process is influenced by external stimuli and experiences. The most well-known mechanism for synaptic plasticity is long-term potentiation (LTP) which is a process where the strength of a synapse is enhanced following repeated activation of the presynaptic neuron.

However, the human brain is significantly more complex than existing artificial neural networks and spiking neural networks. Hence, directly reprogramming the weights of human neurons is not currently feasible or practical. However, the principles of neural network training and the use of artificial neural networks and spiking neural networks can provide insights into potential approaches for addressing neurological disorders such as Schizophrenia, Autism, or

**Table 2**
A Table that concludes all achieved experiments

| Scenario | Training Algorithm | Activation Function | Impact of weight loss |
|---|---|---|---|
| 1 | ANN with Backpropagation | Sigmoid | Minor impact on behavior if lost weights are minor and training is still in its initial phases, significant impact if lost weights are significant or training is in later stages. |
| 2 | ANN with Backpropagation | ReLU | Severe impact on behavior and its neighbors, may result in vanishing gradients and make training difficult or impossible. |
| 3 | ANN with Backpropagation | LeakyReLU | Moderate impact on behavior and its neighbors, compensation by neighboring neurons to some extent, but the degree of compensation will depend on the specific implementation. |
| 4 | ANN with Adam optimizer | Sigmoid | Minor impact on behavior and its neighbors. Adapt to change and update weights. The output of a neuron is continuous. Similarities to a human brain: Adaptive learning rates, momentum |
| 5 | SNN with Backpropagation | Sigmoid | Complex impact on behavior and its neighbors. Reduction in output firing rate. The output of a neuron is Binary (spike or no spike). |
| 6 | SNN with SUR algorithm | SReLU | Complex impact on behavior and its neighbors. Reduction in output firing rate. The output of a neuron is Binary (spike or no spike). Similarities to a human brain: Synaptic plasticity. |

Alzheimer's disease. One approach that has shown promise is deep brain stimulation (DBS), which involves implanting electrodes in specific areas of the brain and delivering electrical impulses to regulate neural activity. DBS has been used successfully to treat a variety of neurological and psychiatric disorders, including Parkinson's disease, depression, and obsessive-compulsive disorder Fins, Mayberg, Nuttin et al. (2011). Another approach is to develop drugs or therapies that target specific neurotransmitters or receptors in the brain. For example, medications that block the activity of the neurotransmitter dopamine have been used to treat symptoms of schizophrenia, while drugs that increase the availability of acetylcholine have been used to improve memory and cognitive function in Alzheimer's patients Kapur and Mamo (2003); Jefferson (2003). In addition, recent research has shown that deep learning algorithms, which are used to train artificial neural networks, can be used to analyze brain imaging data and identify patterns associated with neurological disorders. This approach could potentially lead to new diagnostic tools and personalized treatments for individuals with neurological disorders Orrù, Pettersson-Yeo, Marquand, Sartori and Mechelli (2012); Arbabshirani, Plis, Sui and Calhoun (2017).

### 4.1. Adam optimzer

As noticed in the upper-mentioned section, while there are some parallels between the Adam optimizer and the way the human brain learns, it is important to note that both systems are very different in terms of their complexity and the way they process information. Therefore, it seems so difficult to directly use the Adam optimizer to re-train human neurons for treating specific disorder conditions. However, creating an Adam-based new reinforced learning algorithm that is designed to mimic the way the brain responds to positive and negative feedback can help identify new patterns or relationships in complex data.

The Adam optimizer could be mathematically described by the following equations:
The moving average of the gradient:

$$m_t = \beta_1 m_{t-1} + (1 - \beta_1) g_t \qquad (1)$$

The moving average of the squared gradient:

$$v_t = \beta_2 v_{t-1} + (1 - \beta_2) g_t^2 \qquad (2)$$

The bias correction for the moving average of the gradient:

$$\hat{m}_t = \frac{m_t}{1 - \beta_1^t} \qquad (3)$$

The bias correction for the moving average of the squared gradient:

$$\hat{v}_t = \frac{v_t}{1 - \beta_2^t} \qquad (4)$$

The update rule for the weights:

$$w_{t+1} = w_t - \frac{\alpha \hat{m}_t}{\sqrt{\hat{v}_t} + \epsilon} \qquad (5)$$

where, $g_t$ is the gradient at time $t$, $m_t$ and $v_t$ are the moving averages of the gradient and the squared gradient respectively, $\beta_1$ and $\beta_2$ are the decay rates for the moving averages. Typically, $\beta_1$ is set to 0.9 and $\beta_2$ is set to 0.999. $t$ is the time step, $\alpha$ is the learning rate, $\epsilon$ is a small constant to avoid division by zero, $w_t$ is the updated weights at time $t$.

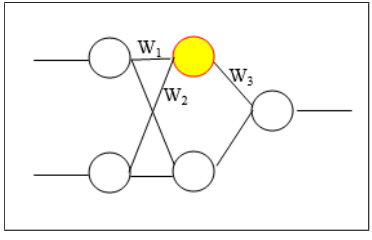

**Figure 3:** A neural network with two inputs, one hidden layer with two units, and one output. The output of the yellow neuron is determined by equation 6.

Even though the Adam optimizer is not a comprehensive model of the biological processes that occur in the brain, the above equations can be conceptually analogized with the way the human brain learns as follows; Kingma and Ba (2014); Lillicrap, Cownden, Tweed and Akerman (2016):

- The moving average of the gradient ($m_t$) can be seen as analogous to the synaptic strength or connection weight between neurons in the brain. Just as the synaptic strength changes based on feedback from the environment, the moving average of the gradient changes based on feedback from the loss function.

- The moving average of the squared gradient ($v_t$) can be seen as analogous to the square of the synaptic strength or connection weight. It provides a measure of the variability or uncertainty in the synaptic strength.

- The bias correction terms ($\hat{m}_t$ and $\hat{v}_t$) adjust the moving averages to account for the initial bias at the beginning of training. Similarly, the brain also adjusts the synaptic strengths to account for any initial biases.

- The update rule for the weights ($w_t$) adjusts the weights based on the moving averages of the gradient and the squared gradient. This is similar to how the synaptic strengths are adjusted based on feedback from the environment in the brain.

Here's an example that illustrates the analogy between the Adam optimizer and the way the human brain learns. Suppose having a neural network with two inputs, one hidden layer with two units, and one output, Figure 3, and we want to train it to predict the output given the inputs. Let's assume that the inputs are scalar values $x_1$ and $x_2$, the output is a scalar value y, and the network has two hidden units with ReLU activation functions. We can write the network as:

$$y = f(w_3 \cdot \max(0, w_1 x_1 + w_2 x_2) + b) \quad (6)$$

where $f$ is the output activation function, $w_1$ and $w_2$ are the weights connecting the input to the hidden units, $w_3$ is the weight connecting the hidden units to the output, and $b$ is the bias term. The $\max(0, x)$ function applies the ReLU activation function to the sum of the weighted inputs to the hidden units.

To train this network, we need to define a loss function that measures the difference between the predicted output and the true output for a given input. Let's use the mean squared error (MSE) as the loss function here:

$$L = \frac{1}{2}(y_{true} - y_{pred})^2 \quad (7)$$

where $y_{true}$ is the true output and $y_{pred}$ is the predicted output.

To compute the moving average of the gradient, the squared gradient, the bias-corrected estimates of the moving averages, and the update to the weights according to the upper-mentioned equations (1, 2, 3, 4, and 5) respectively, we need to calculate the gradient of the loss with respect to the weights:

$$g_t = \nabla_w L(w_t) \quad (8)$$

The weights update rule has some similarities with the way that the human brain adjusts the strength of connections between neurons during learning. In the brain, synaptic plasticity, which is the ability of synapses to change in strength, is thought to be a key mechanism underlying learning and memory. One form of synaptic plasticity is long-term potentiation (LTP), which is a process by which the strength of a synapse is increased following repeated activation of the presynaptic neuron. LTP is thought to involve the activation of certain signaling pathways, such as the NMDA receptor pathway, and the subsequent strengthening of the synapse by the insertion of more receptors or the growth of new dendritic spines.

The NMDA receptor pathway refers to a specific signaling pathway in the brain that involves the N-methyl-D-aspartate (NMDA) receptor, a type of glutamate receptor that is involved in synaptic plasticity and learning and memory processes. When glutamate, a neurotransmitter, binds to the NMDA receptor, it triggers a cascade of events that lead to changes in the strength and structure of the synapse, the junction between two neurons. This process is known as synaptic plasticity and is thought to underlie learning and memory formation. The NMDA receptor pathway involves several key signaling molecules, including calcium ions (Ca2+), protein kinases, and transcription factors, which work together to initiate and maintain changes in synaptic strength. Dysregulation of the NMDA receptor pathway has been implicated in several neurological and psychiatric disorders, including Alzheimer's disease, schizophrenia, and depression. In this context, the Adam optimizer can be seen as a computational model of this process, where the weights of the neural network are the synapses, the gradient of the loss function is the signal that activates the presynaptic neuron, and the update rule is the process by which the strength of the synapse is increased or decreased. Just as in the brain, the Adam optimizer uses a form of "memory" to keep track of the history of the gradients and adjust the update rule accordingly. This allows the optimizer to adapt to the structure of the loss function and converge to a good solution quickly.

It is important to reiterate here that the conceptual analogy drawn between the Adam optimizer and the learning process of the human brain should not be interpreted as a direct comparison between the mathematical equations used in the optimizer and the biological mechanisms that take place in the brain. Rather, this analogy serves as a helpful tool to comprehend the underlying principles of the optimization algorithm. By emphasizing the significance of balancing exploration and exploitation during the learning process, this analogy sheds light on how to achieve efficient and effective learning.

### 4.2. Human brain training; practical and ethical considerations

There is growing evidence that using technology-based interventions, such as virtual reality and computerized cognitive training, can improve outcomes for individuals with neurological disorders. For example, studies have shown that virtual reality-based interventions can improve motor function and reduce pain in individuals with Parkinson's disease and stroke Liao, Wu and Hsieh (2017); Pompeu, Arduini, Botelho, Fonseca and Pompeu (2014). Additionally, computerized cognitive training has been found to improve cognitive function in individuals with traumatic brain injury and multiple sclerosis Goverover, Chiaravalloti, O'Brien, DeLuca and Ehrlich-Jones (2018); Akerlund, Esbjornsson and Sunnerhagen (2013) Furthermore, there is evidence that social support and social interactions can improve outcomes for individuals with neurological disorders. For example, studies have found that social support can improve quality of life and reduce depression in individuals with multiple sclerosis Levin, Hadgkiss, Weiland and Jelinek (2017); Siebert, Siebert and Rees (2014) In this context, incorporating visual and acoustic data from past social and familial experiences into training programs that utilize technology-based interventions (such as reinforcement learning or unsupervised learning techniques) and social support could have potential benefits for individuals with neurological disorders.

However, it is so important to note that the implementation of such an approach would require a lot of careful and ethical considerations. One potential issue with using past experiences to train machine learning models is that it may be difficult to ensure that the data is truly representative of a person's experiences. Memories can be unreliable and subjective, and people may have different interpretations of events that occurred in the past. Therefore, it may be challenging to collect and interpret data in a way that accurately reflects a person's experiences. Additionally, using such data could raise concerns about privacy and informed consent. Furthermore, even if the data is accurately collected and the person's privacy and autonomy are respected, there is still the question of whether machine learning models can truly capture the complexity of human mental disorders. While machine learning models can be powerful tools for analyzing complex data, they are not capable of replicating the full complexity of the human brain and its functions.

## 5. Conclusion

This article emphasizes the significance of bridging the gap between biological neuroscience and artificial neural networks in order to enhance our comprehension and treatment of neurological disorders. It begins by investigating the effect of a non-functioning neuron on the performance of artificial neural networks (ANNs), conducting multiple tests using various training algorithms and activation functions to determine the specific impact of the training process on neighboring neurons and the overall performance of the ANN in such scenarios. The study's results have the potential to improve the functionality of inhibited or damaged neurons in individuals with behavioral disorders in the biological field by suggesting the implementation of multiple training phases that incorporate data from past social and familial experiences, both visual and acoustic, to attain this goal. The article also investigates the conceptual analogy between the Adam optimizer and the learning process of the brain, in spite of the significant difference in terms of complexity and information processing. The development and implementation of these approaches must take into account practical and ethical considerations such as treatment safety and efficacy, access to care, and their impact on individuals' autonomy and privacy.

## Acknowledgement

In order to enhance the depth of this research and advance its findings, the author of this paper is actively seeking a collaborative partner in the field of neuroscience who are interested in working to develop an algorithmic or mathematical model that effectively describes the chemical activation function that could be observed in biological neurons.

## References

Akerlund, E., Esbjornsson, E., Sunnerhagen, K.S., 2013. Combined computerized cognitive and physical training in patients with chronic stroke: A pilot study, in: 2013 IEEE-EMBS Conference on Biomedical Engineering and Sciences (IECBES), IEEE. pp. 104–109. doi:10.1109/IECBES.2013.6679498.

Alotaiby, T.N., Malarout, N., 2018. An artificial neural network approach for autism spectrum disorder classification based on the autism diagnostic interview-revised. Journal of Medical Systems 42, 62.

Arbabshirani, M.R., Plis, S., Sui, J., Calhoun, V.D., 2017. Single subject prediction of brain disorders in neuroimaging: Promises and pitfalls. NeuroImage 145, 137–165.

Baba, A., 2021. Advanced ai-based techniques to predict daily energy consumption: A case study. Expert Systems with Applications 184, 115508. doi:10.1016/j.eswa.2021.115508.

Baba, A., 2022. Electricity-consuming forecasting by using a self-tuned ann-based adaptable predictor. Electric Power Systems Research doi:10.1016/j.epsr.2021.107523.

Bengio, Y., Courville, A., Vincent, P., 2015. Representation learning: A review and new perspectives. IEEE Transactions on Pattern Analysis and Machine Intelligence 35, 1798–1828.

Casanova, M.F., Gustafsson, L., Paplinski, A.P., 2004. Neural network modelling of autism, in: Proceedings of the IEEE International Joint Conference on Neural Networks.

Clevert, D.A., Unterthiner, T., Hochreiter, S., 2015. Fast and accurate deep network learning by exponential linear units (elus), in: arXiv preprint arXiv:1511.07289.

Davies, M., Srinivasa, N., Lin, T.H., Chinya, G., Cao, Y., Choday, S.H., et al., 2018. Loihi: A neuromorphic manycore processor with on-chip learning. IEEE Micro 38, 82–99.

Ding, Y., et al., 2020. A deep learning model to predict a diagnosis of alzheimer disease by using 18f-fdg pet of the brain, in: Proc. Radiology, pp. 381–388.

Fins, J.J., Mayberg, H.S., Nuttin, B., et al., 2011. Deep brain stimulation: Ethical issues such as risk perception and deception. Neurosurgery 68, 1–10.

Gerstner, W., Kistler, W.M., Naud, R., Paninski, L., 2014. Neuronal Dynamics: From Single Neurons to Networks and Models of Cognition. Cambridge University Press.

Glorot, X., Bordes, A., Bengio, Y., 2011. Deep sparse rectifier neural networks, in: Proceedings of the Fourteenth International Conference on Artificial Intelligence and Statistics.

Goverover, Y., Chiaravalloti, N.D., O'Brien, A.R., DeLuca, J., Ehrlich-Jones, L., 2018. Computerized cognitive training for individuals with multiple sclerosis: A systematic review and meta-analysis, in: 2018 IEEE 15th International Conference on Rehabilitation Robotics (ICORR), IEEE. pp. 1159–1164. doi:10.1109/ICORR.2018.8460991.

Hoffman, R.E., McGlashan, T.H., 1997. Synaptic elimination, neurodevelopment, and the mechanism of hallucinated "voices" in schizophrenia. The American Journal of Psychiatry 154, 1683–1689.

Jefferson, J.W., 2003. Treatment of alzheimer's disease. Current Psychiatry Reports 5, 9–16.

Kapur, S., Mamo, D., 2003. Half a century of antipsychotics and still a central role for dopamine d2 receptors. Progress in Neuro-Psychopharmacology and Biological Psychiatry 27, 1081–1090.

Kingma, D.P., Ba, J., 2014. Adam: A method for stochastic optimization. arXiv preprint arXiv:1412.6980 .

Kingma, D.P., Ba, J., 2015. Adam: A method for stochastic optimization, in: Proceedings of the IEEE International Conference on Learning Representations.

Lanillos, P., Oliva, D., Philippsen, A.K., Yamashita, Y., Nagai, Y., Cheng, G., 2020. A review on neural network models of schizophrenia and autism spectrum disorder. Neural Networks 122, 338–363.

Levin, A.B., Hadgkiss, E.J., Weiland, T.J., Jelinek, G.A., 2017. Social support and its association with physical and mental health in multiple sclerosis patients: A scoping review, in: 2017 IEEE 17th International Conference on Bioinformatics and Bioengineering (BIBE), IEEE. pp. 190–195. doi:10.1109/BIBE.2017.00040.

Liao, W., Wu, C., Hsieh, Y., 2017. Virtual reality for stroke rehabilitation: A review, in: 2017 International Conference on Applied System Innovation (ICASI), IEEE. pp. 264–267.

Lillicrap, T.P., Cownden, D., Tweed, D.B., Akerman, C.J., 2016. Random feedback weights support learning in deep neural networks. Nature 533, 407–410.

Maass, W., 1997a. Networks of spiking neurons: The third generation of neural network models. Neural Networks 10, 1659–1671.

Maass, W., 1997b. Networks of spiking neurons: The third generation of neural network models. Neural Networks 10, 1659–1671.

Maass, W., 2002. Real-time computing without stable states: A new framework for neural computation based on perturbations. Neural Computation 14, 2531–2560.

Moradi, S., Qiao, N., Stefanini, F., 2017. A scalable multicore architecture with heterogeneous memory structures for dynamic large-scale spiking neural networks. IEEE Transactions on Biomedical Circuits and Systems 11, 128–142.

Orrù, G., Pettersson-Yeo, W., Marquand, A.F., Sartori, G., Mechelli, A., 2012. Using support vector machine to identify imaging biomarkers of neurological and psychiatric disease: A critical review. Neuroscience and Biobehavioral Reviews 36, 1140–1152.

Pfeiffer, M., Brette, R., 2018. Deep learning with spiking neurons: Opportunities and challenges. Frontiers in Neuroscience 12, 774.

Pompeu, J.E., Arduini, L.A., Botelho, A.R., Fonseca, M.B., Pompeu, S.M., 2014. Feasibility, safety and outcomes of playing kinect adventures!™ for people with parkinson's disease: A pilot study, in: 2014 IEEE-EMBS International Conference on Biomedical and Health Informatics (BHI), IEEE. pp. 310–313.

Ruppin, E., Reggia, J.A., Horn, D., 1996. Pathogenesis of schizophrenic delusions and hallucinations: A neural model. Schizophrenia Bulletin 22, 105–123.

Siebert, D.C., Siebert, C.F., Rees, J.H., 2014. Social support and depression in multiple sclerosis: A systematic review, in: 2014 IEEE-EMBS International Conference on Biomedical and Health Informatics (BHI), IEEE. pp. 163–166. doi:10.1109/BHI.2014.6864303.

Srinivasa, N., Cruz-Albrecht, J.M., Chakradhar, S., Cauwenberghs, G., 2016. Spinnaker: Enabling real-time large-scale neural simulations. Proceedings of the IEEE 104, 1023–1037.

Verstraeten, D., Schrauwen, B., D'Haene, M., Stroobandt, D., Campenhout, J.V., 2006. An experimental unification of reservoir computing methods, in: Neural Networks, pp. 173–192.

Wang, R., Li, X., Chen, Y., Xu, J., 2020. A survey on spiking neural networks: Models, learning mechanisms and applications. Neurocomputing 399, 68–88.